\documentclass{article}
\usepackage{ijcai23}

\usepackage{times}
\usepackage{soul}
\usepackage{url}
\usepackage[hidelinks]{hyperref}
\usepackage[utf8]{inputenc}
\usepackage[small]{caption}
\usepackage{graphicx}
\usepackage{amsmath}
\usepackage{amsthm}
\usepackage{booktabs}
\usepackage{algorithm}
\usepackage{algorithmic}
\usepackage[switch]{lineno}

\usepackage{amsfonts,amssymb}
\usepackage{stfloats}
\usepackage{subfigure}
\usepackage{multirow}
\usepackage{color}
\usepackage{gensymb}
\usepackage{hyperref}
\usepackage{cleveref}

\title{Video Frame Interpolation with Densely Queried Bilateral Correlation}

\author{
Chang Zhou
\and
Jie Liu\thanks{Corresponding author.}\and
Jie Tang\And
Gangshan Wu
\affiliations
State Key Laboratory for Novel Software Technology, Nanjing University, China
\emails
zhouchang@smail.nju.edu.cn,
\{liujie, tangjie, gswu\}@nju.edu.cn
}

\begin{document}

\maketitle

\begin{abstract}
   Video Frame Interpolation (VFI) aims to synthesize non-existent intermediate frames between existent frames. Flow-based VFI algorithms estimate intermediate motion fields to warp the existent frames. Real-world motions' complexity and the reference frame's absence make motion estimation challenging. Many state-of-the-art approaches explicitly model the correlations between two neighboring frames for more accurate motion estimation. In common approaches, the receptive field of correlation modeling at higher resolution depends on the motion fields estimated beforehand. Such receptive field dependency makes common motion estimation approaches poor at coping with small and fast-moving objects. To better model correlations and to produce more accurate motion fields, we propose the Densely Queried Bilateral Correlation (DQBC) that gets rid of the receptive field dependency problem and thus is more friendly to small and fast-moving objects. The motion fields generated with the help of DQBC are further refined and up-sampled with context features. After the motion fields are fixed, a CNN-based SynthNet synthesizes the final interpolated frame. Experiments show that our approach enjoys higher accuracy and less inference time than the state-of-the-art. Source code is available at https://github.com/kinoud/DQBC.
\end{abstract}

\section{Introduction}

Video Frame Interpolation (VFI) aims to synthesize non-existent intermediate frames between existent frames. VFI is used for a wide range of applications such as slow-motion generation \cite{superslomo}, video compression \cite{compression}, video restoration \cite{restore1,restore2,restore3}, novel view synthesis \cite{novel_view,novel_view2}, animation production \cite{animeinterp,anime2,anime3} and video prediction \cite{vfp}. Nowadays, the flow-based paradigm dominates the methodology of VFI owing to its superior performance. However, the absence of the reference frame makes it challenging to estimate the intermediate optical flow fields or motion fields used to warp the neighboring frames backward. \cite{dain,superslomo,liu2019deep,context_aware,toflow} approximate the intermediate motion fields by linearly combining the optical flow fields between the two input frames. \cite{film,rife,ifrnet} directly estimate the motion fields using Convolutional Neural Networks (CNN) in a coarse-to-fine manner. \cite{bmbc,abme,vfiformer} explicitly model correlations between two input frames, which appears as a strong prior in the VFI task and helps to produce more accurate motion fields.

\begin{figure}[tbp]
    \centering
    \includegraphics[width=0.85\linewidth]{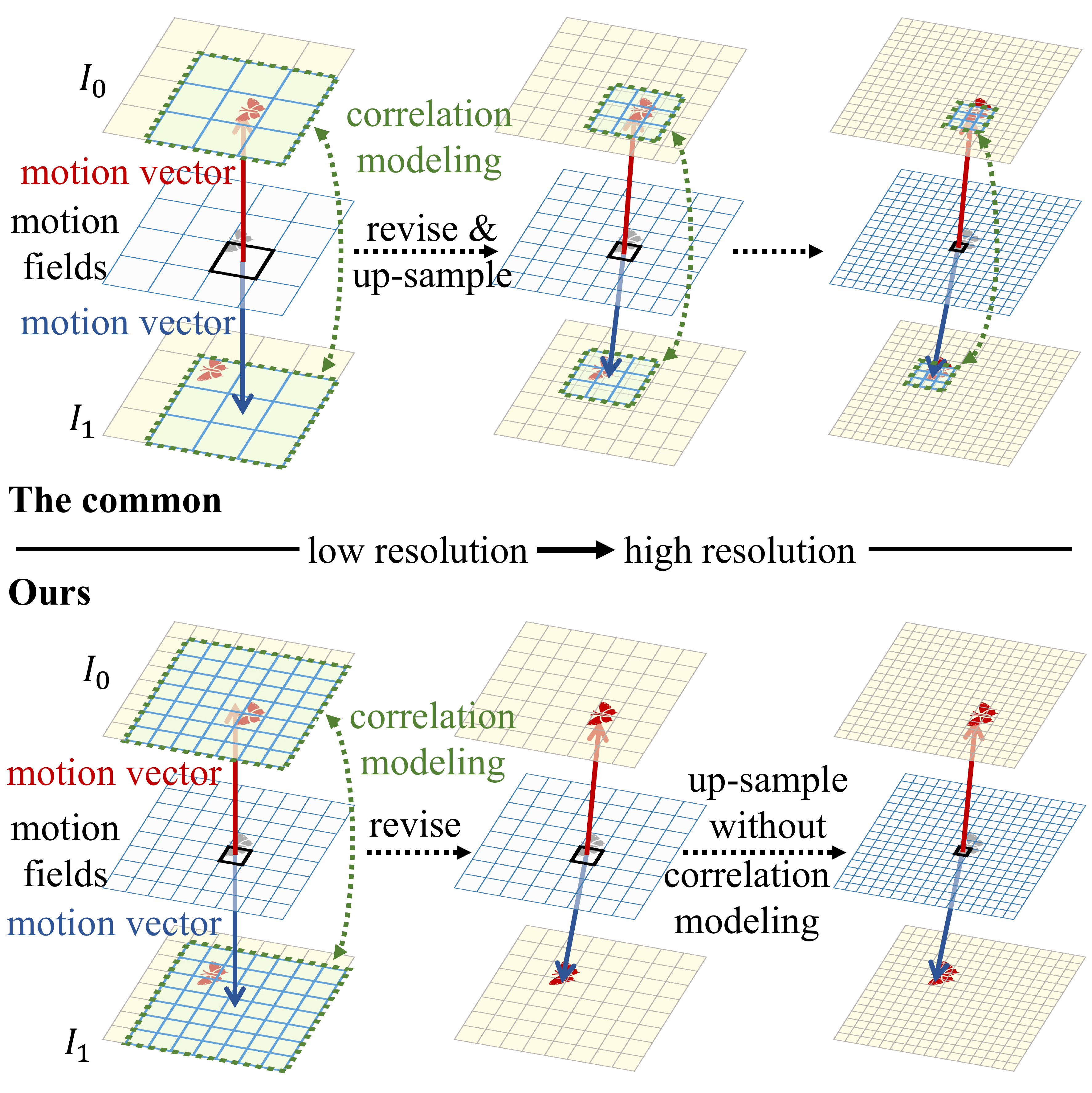}
    \caption{The common correlation modeling paradigm \textit{vs.} ours. The correlation modeling is conducted along with the motion estimation process. The common approach iteratively models correlations from low resolution to high resolution. Our approach models correlations at a single resolution. } 
    \label{fig:summary}
\end{figure}

Correlation modeling is often conducted locally in a coarse-to-fine manner coupled with the motion fields estimation as in \cite{bmbc,abme,vfiformer}. The top half (``The common") of \Cref{fig:summary} illustrates this idea. At each resolution level, for every location in the motion fields, two local windows on $I_0$ and $I_1$ are picked to model correlations between them. In correlation modeling, similarities of patches in two local windows (Each local window contains $3\times 3$ patches in the top half of \Cref{fig:summary}.) are computed. These similarity values play an important role in motion fields revisions performed by neural networks. After being revised, the motion fields are up-sampled, and correlation modeling is conducted again at a higher resolution. This process is repeated across several resolution levels.

However, such a paradigm suffers from the ``receptive field dependency" problem, which will be explained in detail next. As shown in the top half of \Cref{fig:summary} , correlation modeling is conducted within two local windows, and the locations of the two local windows are determined by motion vectors. Considering the local windows are relatively small at higher resolution, if the motion vectors estimated at low resolution are inaccurate, the local windows at higher resolution may not cover the butterfly at all. In that case, the correlation modeling at higher resolution does not work as expected because the local windows are not placed in the intended areas. In other words, the receptive field of correlation modeling depends on the motion fields estimated beforehand. So, in the common approach shown in the top half of \Cref{fig:summary}, we expect that the motion fields estimated at lower resolution are reasonably accurate to prevent misleading the correlation modeling at higher resolution. However, it is not fair to expect the motion fields estimated at low resolution to be ``reasonably accurate", especially for small objects. The reasons are as follows: At low resolution, patches used for computing similarities are relatively large, and small objects may only occupy a small fraction of the patch. Thus the enormous background may dominate the similarity computing regarding patches that contain small objects. Besides, suppose a large patch on one frame includes two or more small objects, and these small objects are separated across different patches on the other frame. In that case, the correlation modeling and motion estimation at low resolution may face ambiguity.

Low-resolution correlation modeling is expected to cope with fast-moving objects but not small objects, and high-resolution correlation modeling is expected to cope with small objects but not fast-moving objects. Superficially, they are complementary. But considering the aforementioned receptive field dependency problem, if low-resolution correlation modeling fails to capture small and fast-moving objects (which may easily happen), high-resolution correlation modeling may also miss that object. The receptive field dependency problem makes the common approach poor at coping with small and fast-moving objects.

In this work, we propose the Densely Queried Bilateral Correlation (DQBC) to better model correlations between two frames and produce more accurate motion fields. We conduct correlation modeling differently compared with the common approach. The bottom half of \Cref{fig:summary} illustrates our paradigm. Instead of modeling correlations iteratively from very low resolution, we model correlations all at once at a single and relatively high resolution. Within the local windows, DQBC is calculated asymmetrically: When computing similarities between patches, query patches keep their resolution and key patches are down-sampled several times for larger receptive fields. DQBC gets rid of the receptive field dependency problem and helps generate more accurate motion fields, especially for small and fast-moving objects. To refine and up-sample the motion fields to full size, we further devise a Motion Refinement Module (MRM). MRM refines and up-samples the input motion fields with the help of context features. Compared with \cite{vfiformer} that uses heavy transformers to synthesize the intermediate frame after motion estimation, we devise a relatively lightweight CNN called SynthNet to synthesize the final result. 

Our contributions are summarized as follows:
\begin{itemize}
    \item We propose the Densely Queried Bilateral Correlation (DQBC) to better model correlations for the VFI task. DQBC gets rid of the receptive field dependency problem and thus is more friendly to small and fast-moving objects.
    \item With the DQBC, our VFI framework achieves state-of-the-art performance on various VFI benchmarks. In our VFI framework, we devise a Motion Refinement Module (MRM) to refine and up-sample the motion fields and a CNN-based SynthNet to synthesize the interpolated frame. 
\end{itemize}

\section{Related Work}

VFI is a long-standing task, and various approaches have been proposed. 

Kernel-based approaches such as \cite{ada_conv,ada_sep_conv,cheng2020video,cheng2021multiple,adacof} estimate spatially-adaptive convolutional kernels for each pixel to be synthesized. \cite{phase_based,phasenet} represent motions in the phase shifts of individual pixels. \cite{cain,dynamic_arch} distribute spatial information into the channel dimension and perform channel attention for pixel-level frame synthesis.

Besides these, the flow-based approaches enjoy a long history and have been prevailing. In recent flow-based VFI works, \cite{context_aware} proposes to warp the context features using the motion fields for better interpolation quality. \cite{superslomo} proposes a network for variable-length multi-frame video interpolation. \cite{dain} exploits the additional depth information to better handle occlusions in the sampling process. \cite{im_net,xvfi} explore high-resolution video interpolation. \cite{time_lens} explores the event-based approach. \cite{bmbc} proposes to model correlations between two frames explicitly via cost volume and is improved by \cite{abme} who further models the asymmetric motions. \cite{rife,ifrnet} aim at efficient VFI algorithms and yet have shown compelling performance. \cite{vfiformer} proposes a transformer-based synthesis network to blend two warped frames better and hallucinate details.

Our work follows the flow-based paradigm and gives focuses on correlation modeling. Among the approaches mentioned above, \cite{bmbc,abme,vfiformer} explicitly model the correlations between two frames. \cite{bmbc} computes similarities between any two centrosymmetric patch pairs between two local windows. \cite{abme} further models the correlation between the intermediate anchor frame produced by \cite{bmbc} and two input frames. \cite{vfiformer} adopts the IFNet \cite{rife} without explicit correlation modeling as the base motion estimator and further refines the motion fields with explicit correlation modeling. It computes similarities between the center patch of one local window and all patches of the other. Despite the difference in the modeling details, they all model correlations locally in a coarse-to-fine manner, as illustrated in the top half of \Cref{fig:summary}.

Our work takes two frames as inputs. There are also VFI works that take four input frames and are beyond the scope of this paper. We refer interested readers to \cite{qvi,eqvi,inp4_1,inp4_2,inp4_3,inp4_4} for more details.

\begin{figure*}[t]
    \centering
    \includegraphics[width=0.9\textwidth]{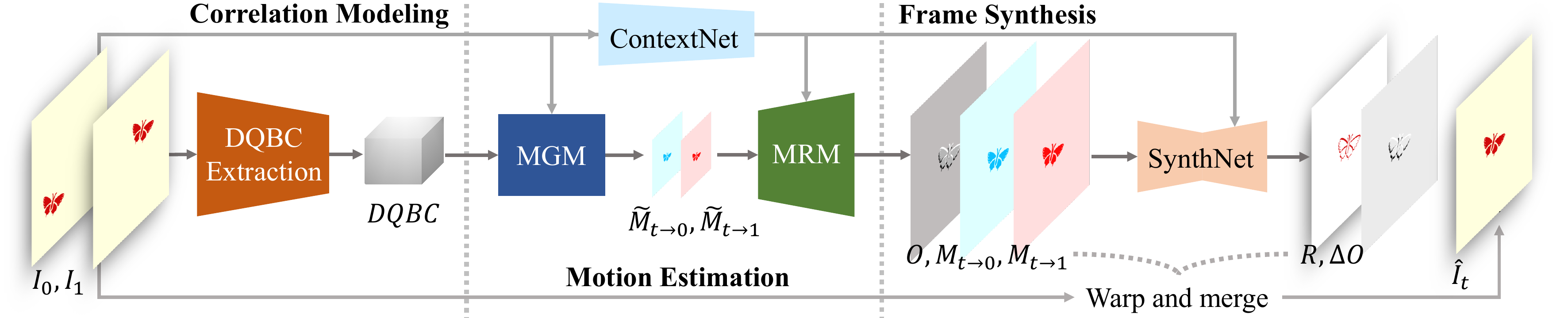}
    \caption{Overview of our VFI pipeline. Firstly, the Densely Queried Bilateral Correlation ($DQBC$) is extracted. The Motion Generation Module (MGM) generates preliminary motion fields with the help of $DQBC$. The Motion Refinement Module (MRM) refines and up-samples the motion fields to full size and produces a preliminary occlusion map $O$. The SynthNet estimates a residual occlusion map $\Delta O$ and a residual image $R$ for final synthesizing.}
    \label{fig:framework}
\end{figure*}

\begin{figure*}[t]
    \centering
    \subfigure[]{
        \includegraphics[width=0.63\textwidth]{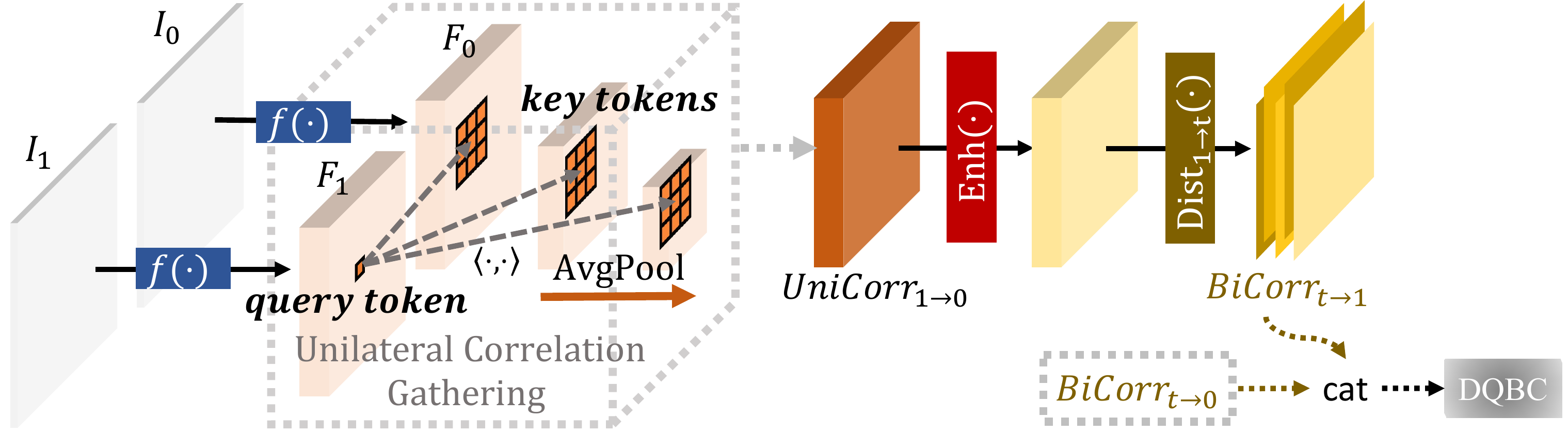}
        \label{fig:dqbc_a}
    }
    \quad
    \subfigure[]{
        \includegraphics[width=0.18\textwidth]{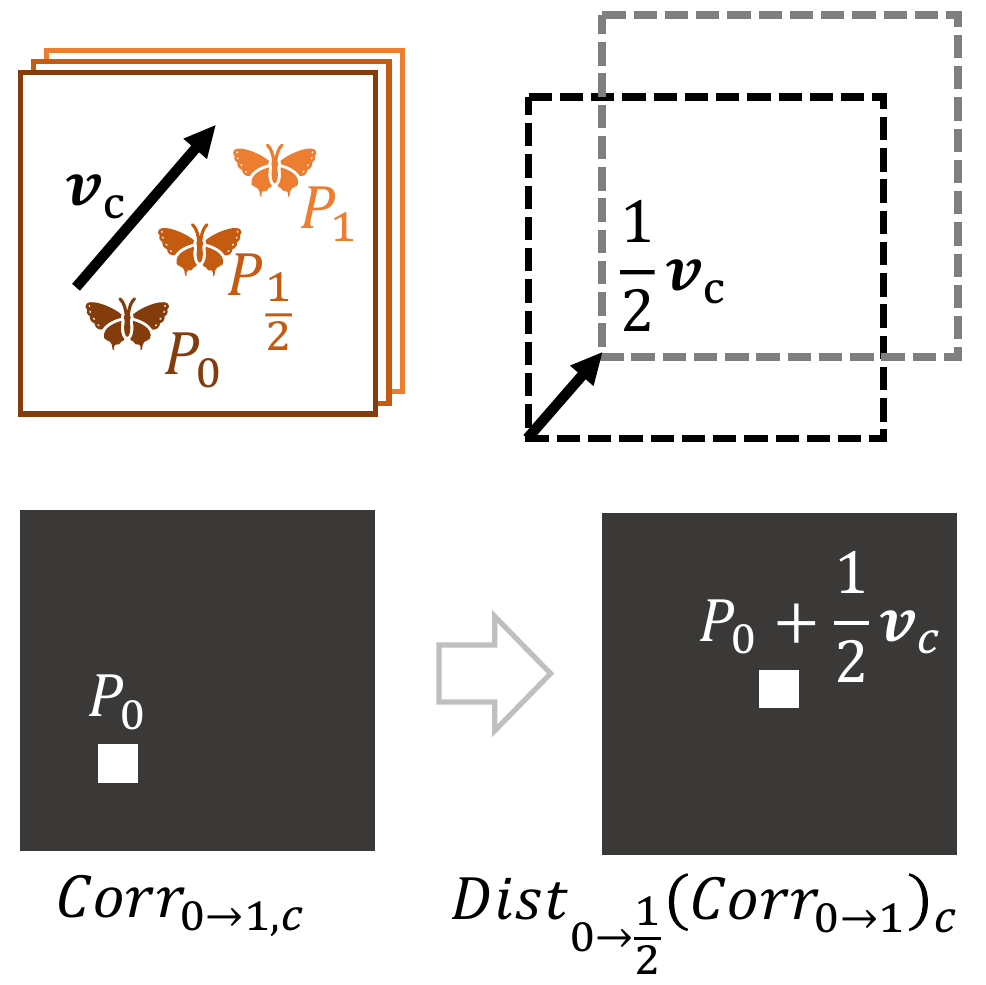}
        \label{fig:dqbc_b}
    }
   \caption{Extraction process of $DQBC$. (a): Tokens in $F_1$ are taken as queries and tokens in local windows on several down-sampled $F_0$ are taken as keys to compute similarities. The unilateral correlation embeddings are then enhanced by the enhancement block and spatially aligned by the feature distributing operation. $DQBC$ is the concatenation of two bilateral correlation embeddings. (b): A toy example illustrating the intuition of spatial alignment. The feature map of unilateral correlation embeddings is spatially aligned with $I_0$ but not with $I_{\frac12}$. To be aligned with $I_{\frac12}$, the entire feature map is shifted by $\frac12\boldsymbol{v}_c$.
    }
    \label{fig:dqbc}
\end{figure*}

\section{The Proposed Method}

\subsection{Overview}

 Given 2 adjacent frames $I_0$ and $I_1$, we synthesize the interpolated frame $\hat{I_t}(t=0.5)$ as:
\begin{equation}
    \hat{I_t} = O\cdot \overleftarrow{w}(I_0,M_{t\xrightarrow{}0})+(\mathbf{1}-O )\cdot \overleftarrow{w}(I_1,M_{t\xrightarrow{}1}) + R.
    \label{eq:syth}
\end{equation}%

In \Cref{eq:syth}, $M_{t\xrightarrow{}\tau}\in \mathbb{R}^{H\times W\times 2}$ is the estimated optical flow field from $\hat{I}_t$ to $I_\tau$ where $\tau\in\{0,1\}$. They are referred to as \textit{motion fields}. $O\in [0,1]^{H\times W}$ is the occlusion map and $R\in \mathbb{R}^{H\times W\times 3}$ is the image residual. The $\overleftarrow{w}(x,y)$ operation is to spatially sample the input tensor $x$ with the vector field $y$.

As shown in \Cref{fig:framework}, our VFI pipeline can be roughly divided into three stages: (1) \textit{Correlation Modeling}, (2) \textit{Motion Estimation}, and (3) \textit{Frame Synthesis}. In stage (1), the Densely Queried Bilateral Correlation ($DQBC$) is extracted from two input frames. In stage (2), $DQBC$ helps to generate preliminary motion fields that are further refined and up-sampled by the Motion Refinement Module (MRM). MRM also produces an initial occlusion map. In stage (3), the SynthNet predicts the image residual and the occlusion map residual. Then the interpolated frame is synthesized following \Cref{eq:syth}.

\subsection{Densely Queried Bilateral Correlation}

The proposed $DQBC$ explicitly encodes the comprehensive correlation information needed for motion estimation.

As shown in \Cref{fig:dqbc_a}, $DQBC$ extraction consists of three parts: (1) \textit{Unilateral Correlation Gathering}, (2) \textit{Correlation Enhancement} (via ``$\text{Enh}(\cdot)"$), (3) \textit{Spatial Alignment} (via ``$\text{Dist}(\cdot)$"), which can be summarized as follows:
\begin{align}
    &BiCorr_{t\xrightarrow{}0} = \mathop{\text{Dist}}\limits_{0\xrightarrow{}t}(\text{Enh}(UniCorr_{0\xrightarrow[]{}1})),\nonumber\\
    &BiCorr_{t\xrightarrow{}1} = \mathop{\text{Dist}}\limits_{1\xrightarrow{}t}(\text{Enh}(UniCorr_{1\xrightarrow[]{}0})),\nonumber\\
    &{DQBC} = [BiCorr_{t\xrightarrow{}0},BiCorr_{t\xrightarrow{}1}],\label{eq:dqbc}
\end{align}
where $[\cdot,\cdot]$ denotes concatenation along channels.

\subsubsection{Unilateral Correlation Gathering}
As shown in \Cref{fig:dqbc_a}, we take one frame as the query frame and the other as the key frame to gather unilateral correlations. Firstly, a feature extractor $f(\cdot)$ extracts high-dimensional visual features $F_0,F_1$ from $I_0,I_1$. $F_1$ is kept as a query set and won't ever be down-sampled. Thus the queries are spatially dense. $F_0$ is down-sampled $(L-1)$ times by factor $1/2$ to form the key sets of different resolution. For every location $(x_q,y_q)$ on $F_1$ and each resolution level $l\in\{0,1,\dots,L-1\}$, we calculate the embedding vector $UniCorr_{1\xrightarrow{}0}^l(x_q,y_q)$ whose $i^\text{th}$ element is:
\begin{align}
    &UniCorr_{1\xrightarrow{}0}^l(x_q,y_q)_i= \nonumber\\
    &\langle F_1(x_q,y_q),F_0^l(\frac{x_q+x_l^i}{2^l},\frac{y_q+y_l^i}{2^l}) \rangle, \label{eq:unicorr}
\end{align}
where $\langle\cdot,\cdot\rangle$ is the inner product operation and $F_0^l$ is spatially down-sampled $F_0$ by factor $1/2^l$ and $(x_l^i,y_l^i)$ is the $i^{\text{th}}$ location in the local window $LW_l$ defined as:
\begin{equation}
    LW_l = \{(2^li,2^lj) | i,j \in \{-r_l,-r_l+1,\dots,r_l\}\}, \label{eq:localwin}
\end{equation}
where $r_l$ is the radius of this local window.

Then the embedding vectors from all resolution levels at $(x_q,y_q)$ are concatenated together to form the unilateral correlation embedding for $(x_q,y_q)$:
\begin{align}
    &UniCorr_{1\xrightarrow{}0}(x_q,y_q)=\nonumber\\
    &\mathop{\text{Concatenate}}\limits_{l\in\{0,1,\dots,L-1\}}\{ UniCorr_{1\xrightarrow{}0}^l(x_q,y_q)\}.\label{eq:int_uc}
\end{align}

Note that $UniCorr_{0\xrightarrow{}1}$ can be acquired in a similar way.

\subsubsection{Correlation Enhancement}

In \Cref{fig:dqbc_a}, the correlation enhancement block $\text{Enh}(\cdot)$ performs denoising and enhancement on unilateral correlation embeddings. Due to occlusions and 3D rotations, some objects may only appear on one of the two input frames. Thus the vanilla unilateral correlation embeddings may contain noise. The $\text{Enh}(\cdot)$ consists of two convolutional layers and a skip connection.

\subsubsection{Spatial Alignment}
The enhanced unilateral correlation embeddings are not spatially aligned with the intermediate frame. In \Cref{fig:dqbc_a}, the feature distributing operation $\text{Dist}(\cdot)$ performs spatial alignment by shifting feature maps.

We first define the displacement vector for each feature map of $UniCorr$. According to \Cref{eq:unicorr}, the $i^\text{th}$ feature map of $UniCorr_{1\xrightarrow{}0}^l$ encodes the similarities between query locations on $I_1$ and key locations on $I_0$. We define the displacement vector as the displacement from the query location to the key location. Thus according to \Cref{eq:unicorr}, the displacement vector $\boldsymbol{v}_i$ of the $i^\text{th}$ feature map of $UniCorr_{1\xrightarrow{}0}^l$ is just $(x_l^i,y_l^i)$. \Cref{eq:int_uc} implies that every feature map of $UniCorr_{1\xrightarrow{}0}$ also has its displacement vector.

The $\text{Dist}(\cdot)$ function in \Cref{eq:dqbc} shifts each feature map of the input embeddings by a fraction of its displacement vector. $\mathop{\text{Dist}}\limits_{0\xrightarrow{}t}(\cdot)$ shifts each feature map of its input by $t\boldsymbol{v}$ and $\mathop{\text{Dist}}\limits_{1\xrightarrow{}t}(\cdot)$ shifts each feature map of its input by $(1-t)\boldsymbol{v}$ where $\boldsymbol{v}$ is the displacement vector of the feature map.

\Cref{fig:dqbc_b} illustrates the intuition of $\text{Dist}(\cdot)$. A butterfly appears at $P_0,P_{\frac12},P_1$ on $I_0,I_{\frac12},I_1$ respectively. Following \Cref{eq:unicorr} and \Cref{eq:int_uc}, we first gather the unilateral correlation embeddings from $I_0$ to $I_1$, denoted as $Corr_{0\xrightarrow[]{}1}$. By definition, the value at $(x,y)$ on the $c^{\text{th}}$ feature map of $Corr_{0\xrightarrow[]{}1}$ (denoted as $Corr_{0\xrightarrow[]{}1,c}$) is the similarity between $(x,y)$ on $I_0$ and $(x,y)+\boldsymbol{v}_c$ on $I_1$ where $\boldsymbol{v}_c$ is the displacement vector of this feature map. If $P_0+\boldsymbol{v}_c$ is reasonably close to $P_1$, $Corr_{0\xrightarrow[]{}1,c}$ will have a relatively high activation value at $P_0$. In $\text{Dist}_{0\xrightarrow{}\frac12}(\cdot)$, we displace this high activation value at $P_0$ to be near $P_{\frac12}$ by shifting the entire feature map by $\frac12\boldsymbol{v}_c$.

Now we have described every part of the $DQBC$ extraction process summarized in \Cref{eq:dqbc}. The $DQBC$ is $1/8$ in spatial resolution of $I_0$ and $I_1$ since the feature extractor $f(\cdot)$ contains three convolutional layers with stride 2.

\subsection{Motion Estimation}

As shown in the middle part of \Cref{fig:framework}, in the motion estimation stage, the Motion Generation Module (MGM) generates preliminary motion fields facilitated by $DQBC$ and then the Motion Refinement Module (MRM) refines and up-samples the motion fields to full size.

\subsubsection{Motion Generation}

With the extracted $DQBC$, MGM generates the preliminary motion fields as:
\begin{align}
    [\tilde{M}_{t\xrightarrow{}0},\tilde{M}_{t\xrightarrow[]{}1}] = \boldsymbol{g}([\boldsymbol{c}([I_0,I_1]),\boldsymbol{m}({DQBC})]).\label{eq:motion_gen}
\end{align}

In \Cref{eq:motion_gen}, $[\cdot,\cdot]$ denotes concatenation along channels. $\boldsymbol{c}(\cdot)$ is a CNN that extracts context information from two input frames. $\boldsymbol{m}(\cdot)$ is a Multi-Layer Perceptron (MLP) with one hidden layer to perform dimensionality reduction on $DQBC$. $\boldsymbol{g}(\cdot)$ is a CNN to generate the motion fields. 

The preliminary motion fields are $1/8$ in spatial resolution of $I_0$ and $I_1$, just the same as $DQBC$. 

\subsubsection{Motion Refinement}

MRM up-samples the motion fields to full size and compensates more details with the warped context features.

Context features are extracted by ContextNet which contains three down-sampling convolutional blocks. It takes $I_\tau$ as input and produces the context feature $C_\tau^i(i\in\{0,1,2\})$ in $1/2^{3-i}$ spatial resolution of $I_\tau$.

MRM consists of three UpBlocks each of which up-samples the input motion fields by factor 2. The first UpBlock takes the motion fields produced by \Cref{eq:motion_gen} and context features $C_0^0,C_1^0$ as inputs and produces the up-sampled motion fields $\tilde{M}_{t\xrightarrow{}0}^1,\tilde{M}_{t\xrightarrow{}1}^1$ and a hidden feature $H_1$ as:
\begin{align}
    &\tilde{M}_{t\xrightarrow{}0}^1,\tilde{M}_{t\xrightarrow{}1}^1,H_1 = \nonumber\\
    &UpBlock_1(\tilde{M}_{t\xrightarrow{}0},\tilde{M}_{t\xrightarrow{}1},C_0^0,C_1^0).
\end{align}

The second and third UpBlock are similar to the first UpBlock but also take the hidden feature produced from the previous UpBlock as input:
\begin{align}
     &\tilde{M}_{t\xrightarrow{}0}^{i+1},\tilde{M}_{t\xrightarrow{}1}^{i+1},H_{i+1} = \nonumber\\
    &UpBlock_{i+1}(\tilde{M}_{t\xrightarrow{}0}^i,\tilde{M}_{t\xrightarrow{}1}^i,C_0^{i},C_1^{i},H_i), \nonumber\\
    &i \in \{1,2\}.
\end{align}

\begin{figure}[hbtp]
    \centering
    \includegraphics[width=0.475\textwidth]{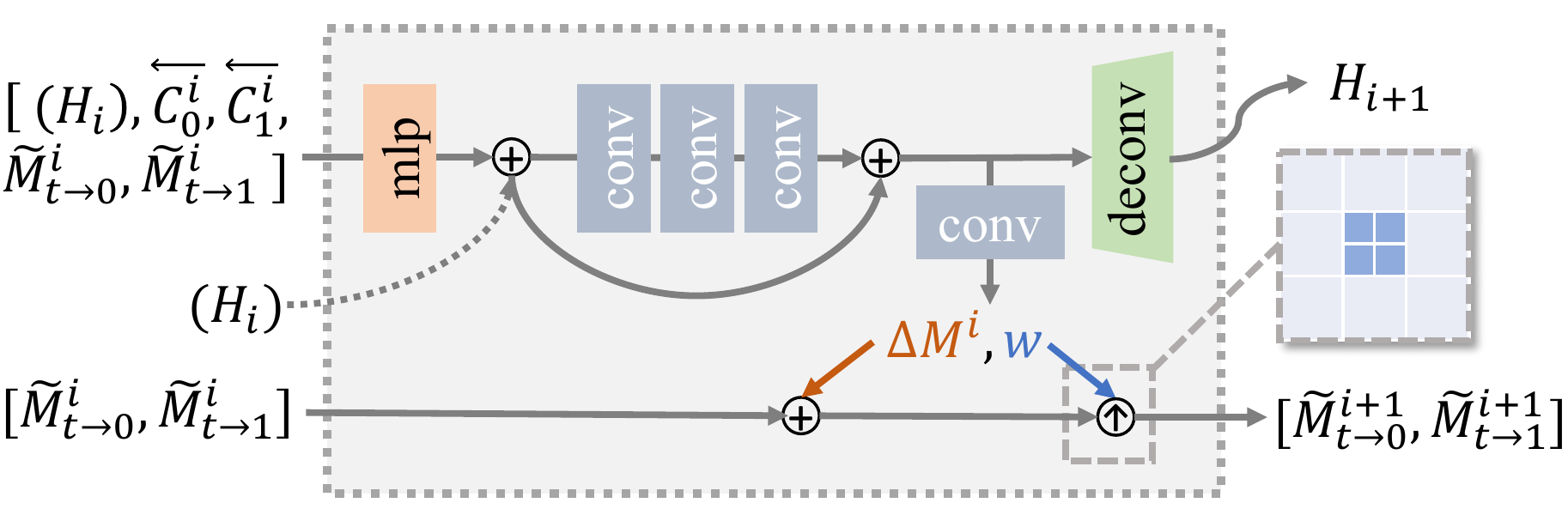}
    \caption{Structure of the UpBlock in the MRM. An Upblock refines and up-samples the input motion fields by a factor of 2.}
    \label{fig:upblock}
\end{figure}

\Cref{fig:upblock} shows the structure of the $(i+1)^\text{th}$ UpBlock ($i\in\{0,1,2\}$). $H_i$ is surrounded by parentheses since there is no $H_0$ for the first UpBlock. $\overleftarrow{C_\tau^i}$ denotes the warped context feature  $\overleftarrow{w}(C_\tau^i,\tilde{M}_{t\xrightarrow{}\tau}^i)$. A convolutional head outputs residual motion fields $\Delta M^i$ and the up-sampling parameter $w$. The input motion fields are added with $\Delta M^i$ and then up-sampled with the parameter $w$. We up-sample the motion fields by taking the convex combination of neighboring $3\times 3$ pixels on the input motion fields. Softmax is performed on $w$ to get the convex combination weights that sum up to 1. 

The occlusion map $O$ is estimated from the last hidden feature $H_3$ using a convolutional head.

\subsection{Frame Synthesis}

We synthesize the interpolated frame following \Cref{eq:syth}. As shown in \Cref{fig:framework}, a SynthNet is devised to estimate a residual occlusion map $\Delta O$ and a residual image $R$ for better interpolation quality. The residual image helps to compensate for details and to eliminate artifacts.

\begin{figure}[h]
    \centering
    \includegraphics[width=0.4\textwidth]{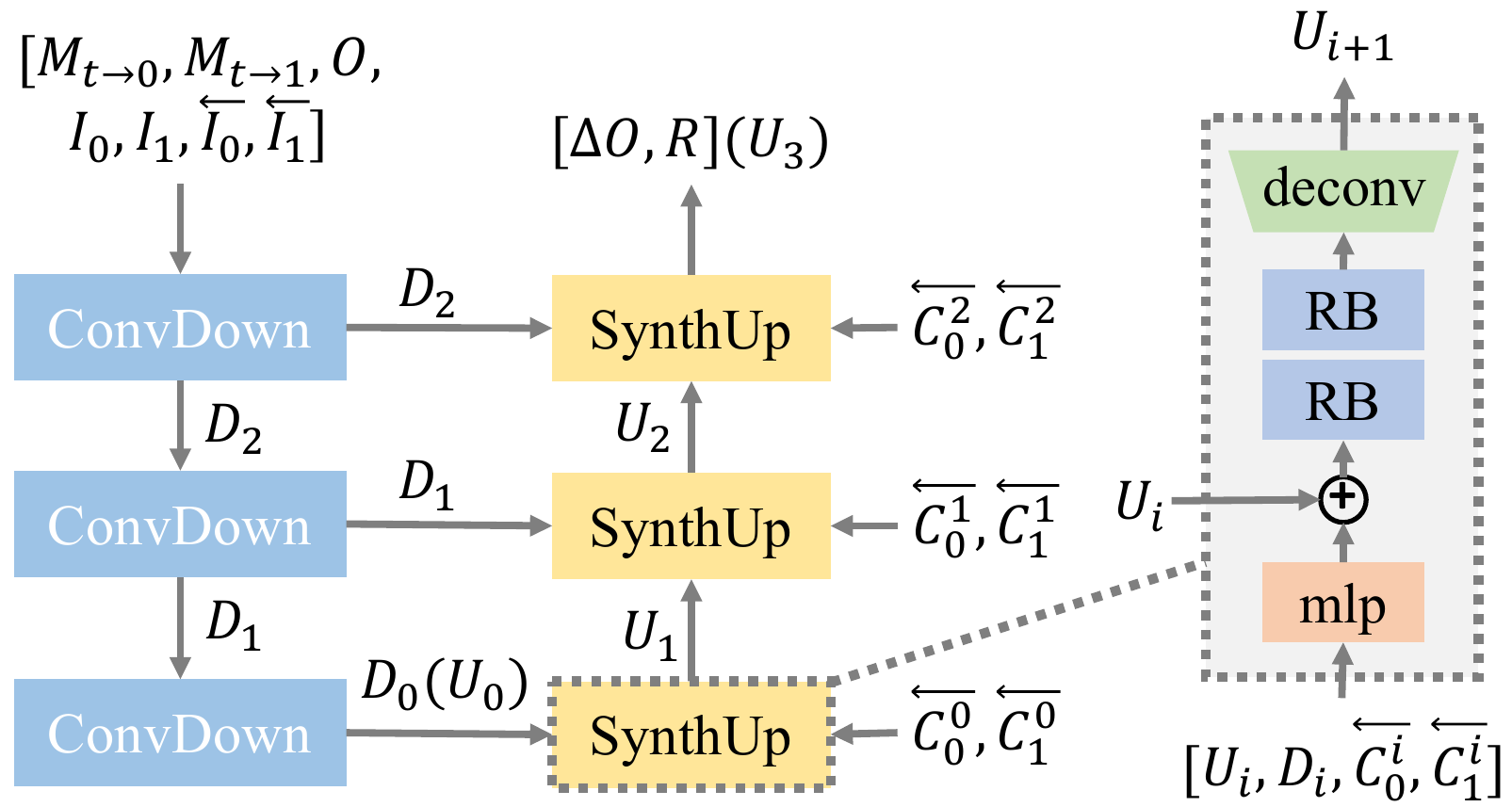}
    \caption{Structure of SynthNet. The Residual Block (RB) consists of 2 convolutional layers and a skip connection.}
    \label{fig:synth}
\end{figure}

Our SynthNet is a modified U-Net as shown in \Cref{fig:synth}. The down-sampling path consists of three ConvDown blocks each of which has two convolutional layers and the first one has stride 2. The up-sampling path consists of three SynthUp blocks that take the warped context features shared by the MRM as auxiliary inputs and fuse them with hidden features.

\subsection{Training}
\subsubsection{Loss Functions}
Our model is trained end-to-end with three loss terms: reconstruction loss $L_{rec}$, teacher reconstruction loss $L_{rec}^{tea}$ and distillation loss $L_{distill}$. The overall loss $L$ is:
\begin{align}
    L = L_{rec} + \lambda_1 L_{rec}^{tea} + \lambda_2 L_{distill}.
\end{align}

$L_{rec}$ is the $L_1$ loss between the estimated intermediate frame $\hat{I}_t$ and the ground truth $I_t^{gt}$:
\begin{align}
    L_{rec} = \Vert \hat{I}_t - I_t^{gt} \Vert_1.
\end{align}

Following \cite{rife}, we adopt a privileged teacher to better supervise the learning of motion estimation. Specifically, during the training process an auxiliary IFBlock \cite{rife} which is accessible to $I_t^{gt}$ takes our estimated motion fields $M_{t\xrightarrow[]{}0},M_{t\xrightarrow[]{}1}$ as preliminary motion fields and produces the revised ones  $M_{t\xrightarrow[]{}0}^{tea},M_{t\xrightarrow[]{}1}^{tea}$. The teacher reconstruction loss is defined as:
\begin{align}
    L_{rec}^{tea} = &\Vert O\cdot \overleftarrow{w}(I_0,M_{t\xrightarrow[]{}0}^{tea})\nonumber \\
    &+ (\boldsymbol{1}-O)\cdot \overleftarrow{w}(I_1,M_{t\xrightarrow[]{}1}^{tea}) -I_t^{gt} \Vert_1,
\end{align}
where $O$ is the occlusion map produced by our model.

The distillation loss is defined as the weighted sum of $L_2$ losses between $M_{t\xrightarrow[]{}0}^{tea},M_{t\xrightarrow[]{}1}^{tea}$ and all the in-process motion fields that our model produces during the motion generation and refinement process.

\subsubsection{Datasets}
We use the training set of the Vimeo90K dataset proposed by \cite{toflow} as our training set. The training dataset contains 51,312 triplets with a resolution of $448\times 256$.

\section{Experiments}

\subsection{Implementation Details}

\subsubsection{Network Achitecture}

The down-sampling levels of key sets when extracting $DQBC$ and the radii ($r_l$ in \Cref{eq:localwin}) of local windows of each resolution level typically determine the preferred motion range that the network copes with. We use three levels of resolution. The radii for each level are 6, 5, and 4, respectively.

\subsubsection{Training Details}

Our network is optimized by AdamW \cite{adamw} with weight decay $10^{-4}$ on $256\times 256$ patches. The patches are randomly cropped from the training set and are randomly flipped, rotated and reversed. We use a total batch size of 64 distributed over 4 Tesla V100 GPUs for 510 epochs. The training process lasts for about three days. The memory consumption for each GPU is about 18 GB. The learning rate gradually decays from $2\times 10^{-4}$ to $2\times 10^{-6}$ using cosine annealing during the training process.

\subsection{Evaluation Datasets}

\paragraph{Vimeo90K.} The evaluation set of Vimeo90K dataset \cite{toflow} contains 3,782 triplets with resolution $448\times 256$.

\paragraph{UCF101.} The UCF101 dataset was originally proposed by \cite{ucf101} and contains human action videos of resolution $256\times 256$. \cite{deepvoxelflow} selected 379 triplets for VFI evaluation.

\paragraph{Middlebury.} Proposed by \cite{middlebury}, the Middlebury benchmark provides two sets: \textit{Other} and \textit{Evaluation}. We adopt the \textit{Other} set which has ground truth intermediate frames and contains 12 triplets for VFI evaluation.

\paragraph{SNU-FILM.} Proposed with \cite{cain}, the SNU-FILM benchmark contains four splits of different difficulty for VFI: \textit{Easy}, \textit{Medium}, \textit{Hard} and \textit{Extreme}. The difficulty is categorized by motion amplitude. Each split contains 310 triplets of resolution up to $1280\times 720$.

\begin{table*}[hbtp]
    \centering
    \resizebox{0.98\linewidth}{!}{
        \begin{tabular}{lcccccccccc}
            \toprule
            \multirow{2}{*}{Method} & 
            \multirow{2}{*}{Vimeo90K} & \multirow{2}{*}{UCF101} & \makebox[0.02\textwidth][c]{Middle-} & \multicolumn{4}{c}{SNU-FILM} & \makebox[0.02\textwidth][c]{Params} & \makebox[0.02\textwidth][c]{MACs} & \makebox[0.01\textwidth][c]{Time}\\
            \cmidrule{5-8}
            & & & bury & Easy & Medium & Hard & Extreme & (M)&(G)&(ms)\\
        
            \midrule
            ToFlow & 33.73/0.9682 & 34.58/0.9667 & 2.15 & 39.08/0.9890 & 34.39/0.9740 & 28.44/0.9180 & 23.39/0.8310 & -&-&-\\
            SepConv & 33.79/0.9702 & 34.78/0.9669 & 2.27 & 39.41/0.9900 & 34.97/0.9762 & 29.36/0.9253 & 24.31/0.8448 & -&-&-\\
            CyclicGen & 32.09/0.9490 & 35.11/0.9684 & - & 37.72/0.9840 & 32.47/0.9554 & 26.95/0.8871 & 22.70/0.8083 & -&-&-\\
            DAIN & 34.71/0.9756 & 34.99/0.9683 & 2.04 & 39.73/0.9902 & 35.46/0.9780 & 30.17/0.9335 & 25.09/0.8584 & -&-&-\\
            CAIN & 34.65/0.9730 & 34.91/0.9690 & 2.28 & 39.89/0.9900 & 35.61/0.9776 & 29.90/0.9292 & 24.78/0.8507 & 42.8&43.5&62\\
            AdaCoF & 34.47/0.9730 & 34.90/0.9680 & 2.24 & 39.80/0.9900 & 35.05/0.9753 & 29.46/0.9244 & 24.31/0.8439 & 21.8 & - & 15\\
            BMBC & 35.01/0.9764 & 35.15/0.9689 & 2.04 & 39.90/0.9902 & 35.31/0.9774 & 29.33/0.9270 & 23.92/0.8432 & 11.0 & 174.2 & 2168\\
            RIFE & 35.61/0.9779 & 35.29/0.9690 & 1.96 & 40.02/0.9904 & 35.72/0.9786 & 30.07/0.9326 & 24.82/0.8529 & 10.1 & 11.7 & 31\\
            ABME & 36.18/0.9805 & 35.38/0.9698 & 2.01 & 39.59/0.9901 & 35.77/0.9789 & 30.58/0.9363 & 25.42/0.8639 & 17.5 & - & 235\\
            VFIformer & \textcolor{blue}{\underline{36.50}}/\textcolor{blue}{\underline{0.9816}} & \textcolor{blue}{\underline{35.43}}/\textcolor{red}{\textbf{0.9700}} & \textcolor{blue}{\underline{1.82}} & 40.13/\textcolor{blue}{\underline{0.9907}} & \textcolor{blue}{36.09}/\textcolor{red}{\textbf{0.9799}} & 30.67/\textcolor{red}{\textbf{0.9378}} & \textcolor{blue}{\underline{25.43}}/\textcolor{blue}{\underline{0.8643}} & 24.2 & 356.7 & 756\\
            \textbf{Ours} & 36.37/0.9812 & 35.35/0.9696 & 1.86 & \textcolor{blue}{\underline{40.15}}/\textcolor{blue}{\underline{0.9907}} & \textcolor{blue}{\underline{36.10}}/0.9796 & \textcolor{blue}{\underline{30.78}}/0.9371 & 25.41/0.8628 & 18.3 & 57.5 & 54\\
            
            \textbf{Ours-Aug} & \textcolor{red}{\textbf{36.57}}/\textcolor{red}{\textbf{0.9817}} & \textcolor{red}{\textbf{35.44}}/\textcolor{red}{\textbf{0.9700}} & \textcolor{red}{\textbf{1.78}} & \textcolor{red}{\textbf{40.31}}/\textcolor{red}{\textbf{0.9909}} & \textcolor{red}{\textbf{36.25}}/\textcolor{red}{\textbf{0.9799}} & \textcolor{red}{\textbf{30.94}}/\textcolor{red}{\textbf{0.9378}} & \textcolor{red}{\textbf{25.61}}/\textcolor{red}{\textbf{0.8648}} & 18.3 & 229.9 & 206\\

            \bottomrule
        \end{tabular}
    }
     \caption{Quantitative comparisons with state-of-the-art VFI methods. The average interpolation error IE (the lower, the better) is reported on the Middlebury benchmark and PSNR/SSIM (the higher, the better) are reported on the others. The \textcolor{red}{\textbf{best}} and \textcolor{blue}{\underline{second-best}} results for accuracy are emphasized with different styles.}
    \label{tab:sota}
\end{table*}

 \subsection{Comparisons with the State-of-the-Art}

We quantitatively compared our model with some other competitive state-of-the-art VFI models, including ToFlow \cite{toflow}, SepConv \cite{ada_sep_conv}, CyclicGen \cite{liu2019deep}, DAIN \cite{dain}, CAIN \cite{cain}, AdaCoF \cite{adacof}, BMBC \cite{bmbc}, RIFE \cite{rife}, ABME \cite{abme} and VFIformer \cite{vfiformer}. Results are presented in \Cref{tab:sota}.

Besides the accuracy metrics (IE/PSNR/SSIM), we also provide \Cref{tab:sota} with numbers of model parameters, MACs and inference time for some methods. MACs are calculated on $256\times 256$ patches using \textit{thop} \footnote{\url{https://pypi.org/project/thop/}}. The inference time is recorded as the average inference time per sample on the Vimeo90K evaluation set and is tested on the same GTX-1080-Ti GPU.

Considering that our base model (``\textbf{Ours}" in \Cref{tab:sota}) runs much faster than VFIformer both theoretically (indicated by MACs) and practically (indicated by inference time), we use test augmentation to boost the performance at the price of more computation. We combine temporal reversal and $90^{\circ}$ spatial rotation to augment the test frame pair to 4 groups and directly average the 4 outcomes to get the final interpolation result. The augmented version is denoted as ``\textbf{Ours-Aug}" in \Cref{tab:sota}.

\begin{figure}[tbp]
    \centering
    \includegraphics[width=0.44\textwidth]{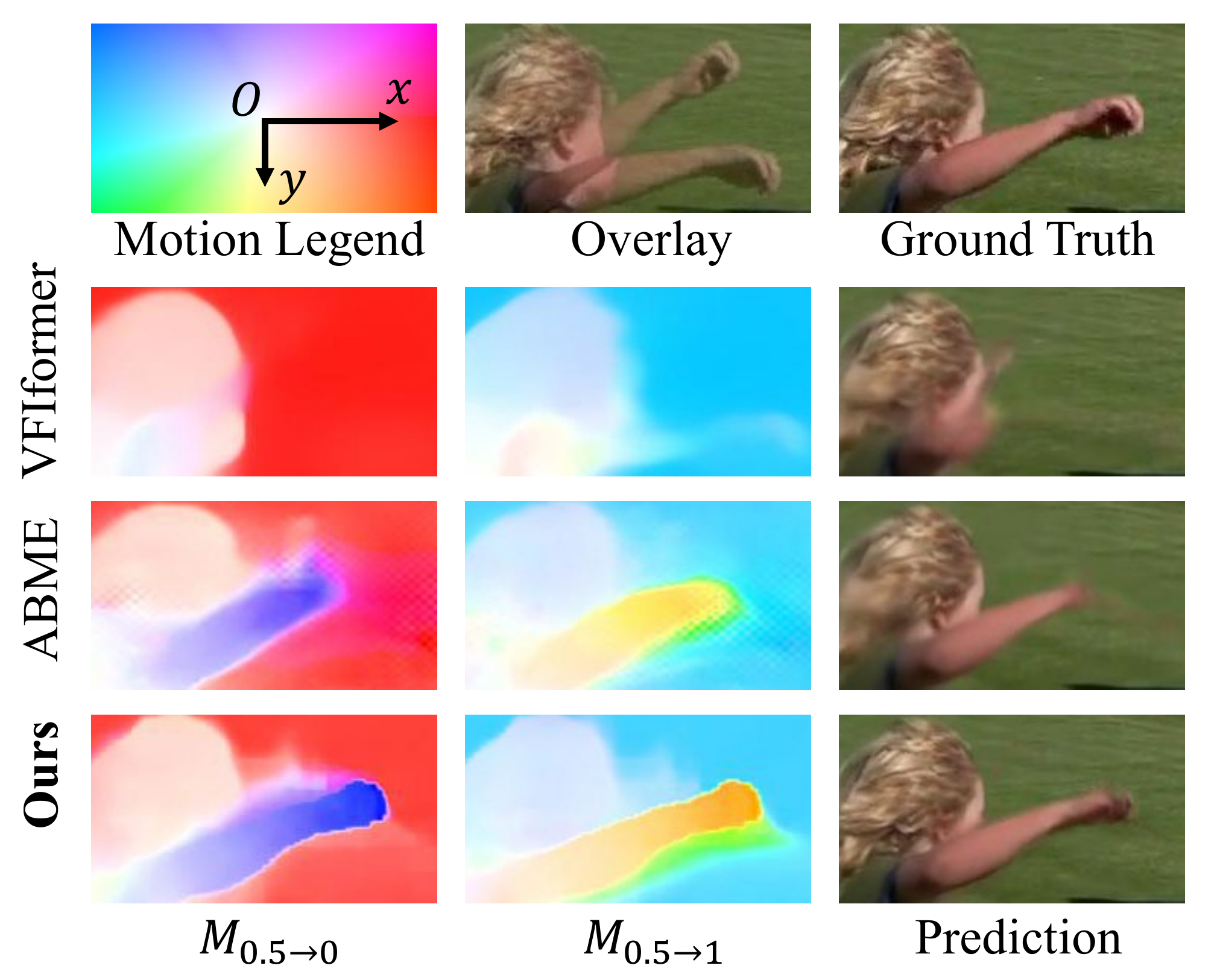}
    \caption{Visualizations of motion fields that are estimated by VFIformer, ABME and DQBC (Ours). Colors represent the directions of motion vectors in motion fields.}
    \label{fig:viz_flow}
\end{figure}

Compared with ABME, our base model achieves \textbf{+0.19} dB improvement and runs in less than \textbf{1/4} inference time on the Vimeo90K evaluation set. Compared with VFIformer, our augmented model achieves \textbf{+0.07} dB improvement and runs in less than \textbf{1/3} inference time on the Vimeo90K evaluation set.

As illustrated by \Cref{fig:viz_flow}, we visualized the motion fields estimated by VFIformer, ABME and our base model. It can be seen that our model is more capable of coping with small and fast-moving objects (the waving arm and fist in this case). The waving arm and fist are totally missed in the motion estimation of VFIformer, leading to a poor interpolation result. ABME succeeds in capturing the motion of the waving arm but misses the fist which is smaller and moves faster.

\begin{figure*}[htp]
    \centering
    \subfigure[Overlay]{
    \begin{minipage}[t]{0.14\linewidth}
    \centering
    \includegraphics[width=\linewidth]{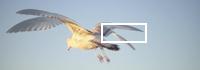}
    \includegraphics[width=\linewidth]{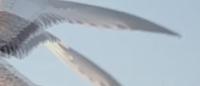}
    \includegraphics[width=\linewidth]{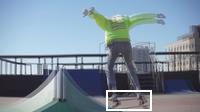}
    \includegraphics[width=\linewidth]{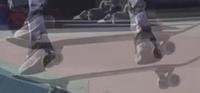}
    \includegraphics[width=\linewidth]{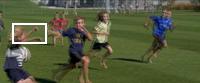}
    \includegraphics[width=\linewidth]{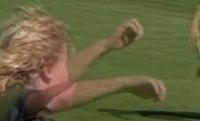}
    \includegraphics[width=\linewidth]{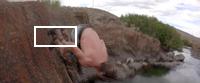}
    \includegraphics[width=\linewidth]{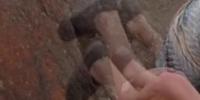}
    \end{minipage}
    }
    \hspace{-3.25mm}
    \subfigure[Ground Truth]{
    \begin{minipage}[t]{0.14\linewidth}
    \centering
    \includegraphics[width=\linewidth]{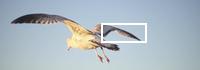}
    \includegraphics[width=\linewidth]{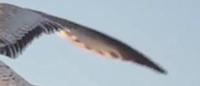}
    \includegraphics[width=\linewidth]{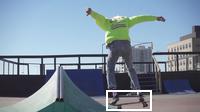}
    \includegraphics[width=\linewidth]{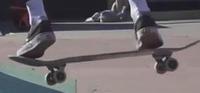}
    \includegraphics[width=\linewidth]{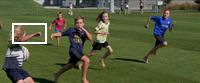}
    \includegraphics[width=\linewidth]{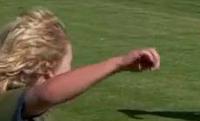}
    \includegraphics[width=\linewidth]{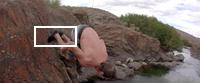}
    \includegraphics[width=\linewidth]{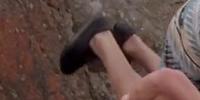}
    \end{minipage}
    }
    \hspace{-3.25mm}
    \subfigure[RIFE]{
    \begin{minipage}[t]{0.14\linewidth}
    \centering
    \includegraphics[width=\linewidth]{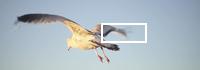}
    \includegraphics[width=\linewidth]{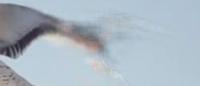}
    \includegraphics[width=\linewidth]{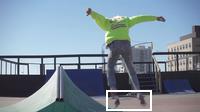}
    \includegraphics[width=\linewidth]{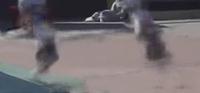}
    \includegraphics[width=\linewidth]{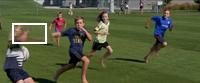}
    \includegraphics[width=\linewidth]{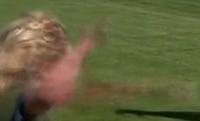}
    \includegraphics[width=\linewidth]{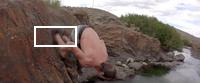}
    \includegraphics[width=\linewidth]{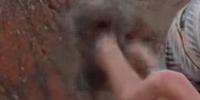}
    \end{minipage}
    }
    \hspace{-3.25mm}
    \subfigure[IFRNet-Large]{
    \begin{minipage}[t]{0.14\linewidth}
    \centering
    \includegraphics[width=\linewidth]{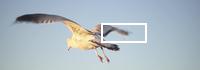}
    \includegraphics[width=\linewidth]{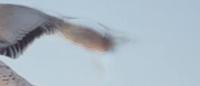}
    \includegraphics[width=\linewidth]{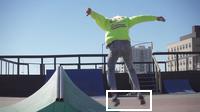}
    \includegraphics[width=\linewidth]{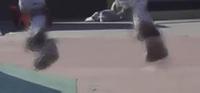}
    \includegraphics[width=\linewidth]{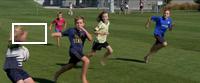}
    \includegraphics[width=\linewidth]{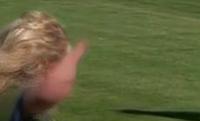}
    \includegraphics[width=\linewidth]{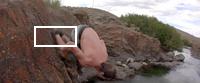}
    \includegraphics[width=\linewidth]{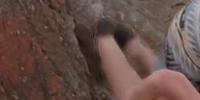}
    \end{minipage}
    }
    \hspace{-3.25mm}
    \subfigure[ABME]{
    \begin{minipage}[t]{0.14\linewidth}
    \centering
    \includegraphics[width=\linewidth]{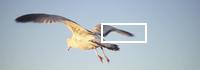}
    \includegraphics[width=\linewidth]{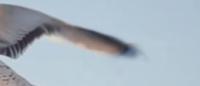}
    \includegraphics[width=\linewidth]{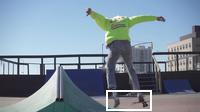}
    \includegraphics[width=\linewidth]{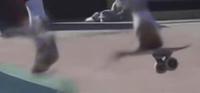}
    \includegraphics[width=\linewidth]{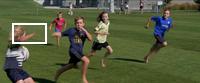}
    \includegraphics[width=\linewidth]{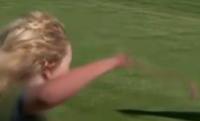}
    \includegraphics[width=\linewidth]{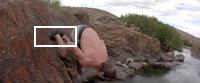}
    \includegraphics[width=\linewidth]{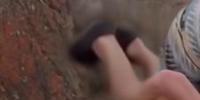}
    \end{minipage}
    }
    \hspace{-3.25mm}
    \subfigure[VFIformer]{
    \begin{minipage}[t]{0.14\linewidth}
    \centering
    \includegraphics[width=\linewidth]{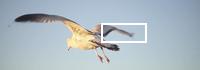}
    \includegraphics[width=\linewidth]{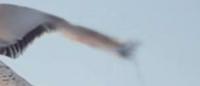}
    \includegraphics[width=\linewidth]{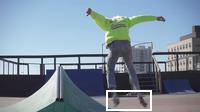}
    \includegraphics[width=\linewidth]{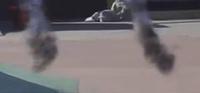}
    \includegraphics[width=\linewidth]{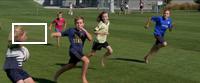}
    \includegraphics[width=\linewidth]{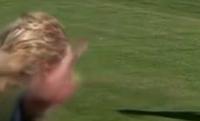}
    \includegraphics[width=\linewidth]{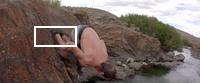}
    \includegraphics[width=\linewidth]{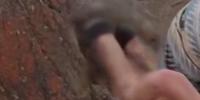}
    \end{minipage}
    }
    \hspace{-3.25mm}
    \subfigure[\textbf{Ours}]{
    \begin{minipage}[t]{0.14\linewidth}
    \centering
    \includegraphics[width=\linewidth]{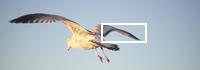}
    \includegraphics[width=\linewidth]{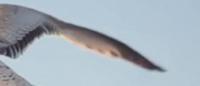}
    \includegraphics[width=\linewidth]{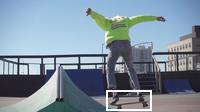}
    \includegraphics[width=\linewidth]{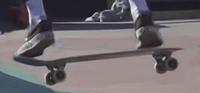}
    \includegraphics[width=\linewidth]{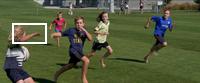}
    \includegraphics[width=\linewidth]{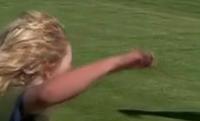}
    \includegraphics[width=\linewidth]{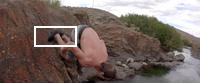}
    \includegraphics[width=\linewidth]{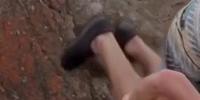}
    \end{minipage}
    }
    
    \caption{Visualizations of interpolated results from some competitive approaches and ours.}
    \label{fig:viz}
\end{figure*}

We also present some visual comparisons of the interpolated results as shown in \Cref{fig:viz}. Both the quantitative comparisons and interpolation visualizations show that our approach is pretty competitive. 

\subsection{Ablation Study}

For the ablation experiments below, we report the results on Vimeo90K evaluation set at 300 training epochs.

\subsubsection{Ablations for DQBC}

To verify the effectiveness of DQBC, we compared our model with the baseline model in which we adopt the bilateral motion estimation module of \cite{abme} as the motion generation module. The motion refinement module and SynthNet stay the same. Basically, the baseline model adopts the paradigm illustrated by the top half of \Cref{fig:summary}. \Cref{tab:ab1} shows that DQBC outperforms the baseline by a notable margin. Besides, it can be seen that the correlation enhancement block and the feature distributing operation also contribute to performance promotion. These results verified the effectiveness of the proposed DQBC.

\begin{table}[htbp]
    \centering
    \setlength{\tabcolsep}{6mm}{
        \begin{tabular}{lcc}
            \toprule
            Model & PSNR & SSIM\\
            \midrule
            Baseline & 36.16 & 0.9804 \\
            Ours  & \textbf{36.31} & \textbf{0.9810} \\
            Ours w/o enh & 36.25 & 0.9809 \\
            Ours w/o dist & 36.28 & 0.9809 \\
            \bottomrule
        \end{tabular}
    }
    \caption{Ablation experiments for DQBC. ``enh" represents the correlation enhancement block and ``dist" represents the feature distributing operation.}
    \label{tab:ab1}
\end{table}

\begin{table}[htbp]
    \centering
    \setlength{\tabcolsep}{5mm}{
        \begin{tabular}{lcc}
            \toprule
            Model & PSNR & SSIM \\
            \midrule
            Bilinear & 36.25 & 0.9808 \\
            Convex w/o refine & 36.11 & 0.9802 \\
            Convex (Ours)  & \textbf{36.31} & \textbf{0.9810} \\
            \bottomrule
        \end{tabular}
    }
    \caption{Contrastive experiments for alternative designs for Motion Refinement Module (MRM).}
    \label{tab:ab2}
\end{table}

\subsubsection{Ablations for Motion Refinement Module}

The Motion Refinement Module (MRM) refines motion fields by producing residuals and up-samples the motion fields by estimating convex combination weights. We compared our approach with the naive bilinear way, where we first bilinearly up-sample the coarse motion fields and then estimate residuals to refine the up-sampled motion fields. We also explored the effectiveness of the refinement function of MRM by removing the residual estimation from it. \Cref{tab:ab2} shows that our approach is better than the naive bilinear approach. The ``convex" approach only works well along with the residual estimation, indicating that it is necessary for MRM to have the ability to revise the motion fields rather than merely up-sample them.

\section{Conclusion}

We have proposed the Densely Queried Bilateral Correlation (DQBC) for better modeling correlations between two input frames in the VFI scenario. DQBC gets rid of the receptive field dependency problem in common approaches and thus is more friendly to small and fast-moving objects. A Motion Refinement Module (MRM) has also been devised to up-sample the motion fields to full size and refine them with context information. We have verified the effectiveness of DQBC and MRM via ablation studies. Both quantitative experiments and interpolation visualizations show that our approach has reached and outperformed the state-of-the-art level on various VFI benchmarks.



\bibliographystyle{named}
\bibliography{vfi}

\end{document}